\documentclass[10pt,twocolumn,letterpaper]{article}
\usepackage[pagenumbers]{cvpr} % To force page numbers, e.g. for an arXiv version
\usepackage[dvipsnames]{xcolor}

\newcommand{\lapapipeline}{%
\begin{figure*}[t]
    \centering
    \includegraphics[width=\textwidth]{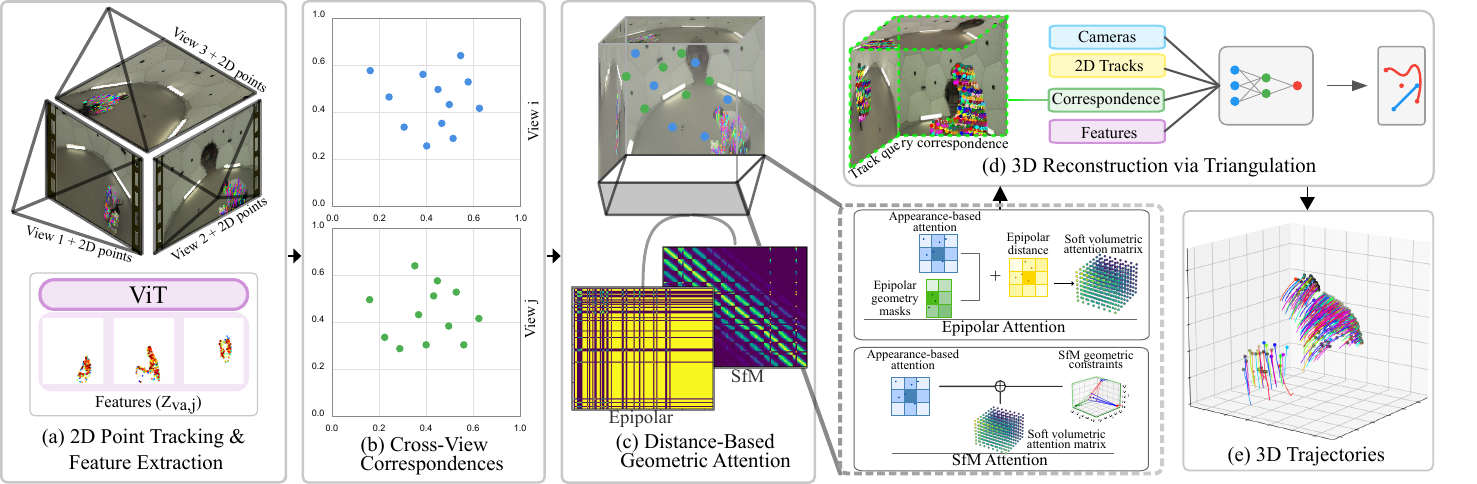}
    \caption{LAPA Architecture Overview: Our end-to-end pipeline processes synchronized multi-view frames through (a) 2D point tracking and feature extraction using Co-Tracker and ViT, (b) cross-view correspondence via volumetric grid creation and distance-based geometric attention, (c) distance-based geometric attention, (d) 3D reconstruction by triangulation using track query correspondence and compound feature integration, and (e) final 3D trajectories with consistent point identities across time and views.}
    \label{fig:lapa-pipeline}
\end{figure*}
}
\newcommand{\volumetricatt}{%
\begin{figure}[t]
    \centering
    \includegraphics[width=0.7\columnwidth]{figures/geometric_attention.pdf}
    \caption{
        Distance-based geometric attention mechanism. 
        We compute attention weights directly from spatial distances $d(G_{v_a}[i], P_{v_a,j})$ rather than feature similarities, 
        using $A_{v_a}(i,j) = \text{softmax}(-d^2/T)$ to establish correspondences between projected grid points and detected 2D points.
    }
    \label{fig:distance-attention}
\end{figure}
}

\newcommand{\qualitativeResults}{%
\begin{figure*}[t]
    \centering
    \includegraphics[width=\textwidth]{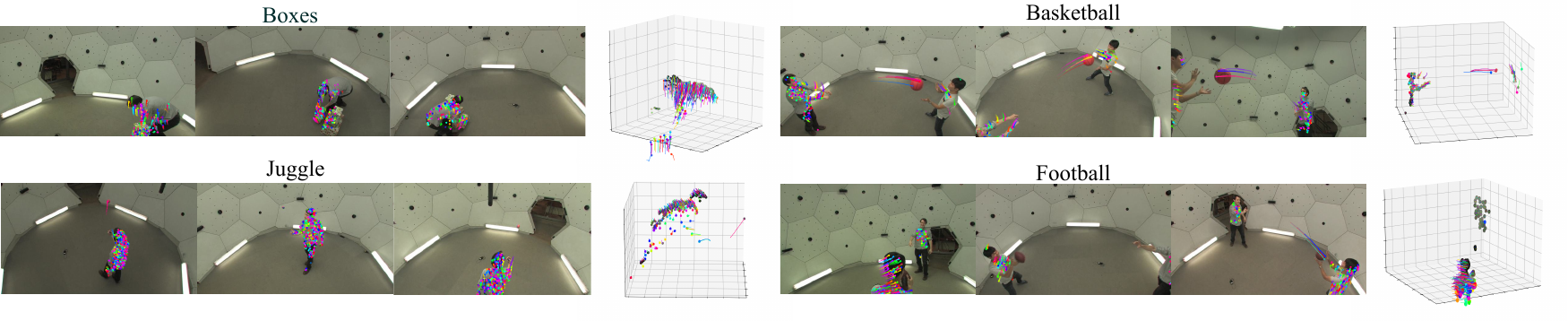}
    \caption{Multi-camera point tracking results demonstrating LAPA's robustness to occlusions and limited camera coverage. We show tracking results on four challenging sequences from TAPVid-3D-MC (Boxes, Juggle, Football, Basketball). Each row presents three synchronized camera views with their corresponding 3D trajectory reconstruction (rightmost). LAPA maintains consistent point identities across all views (shown by consistent colors) by leveraging volumetric attention to aggregate information from all available cameras. The 3D visualizations demonstrate smooth, complete trajectories even when points are occluded or outside individual camera fields of view particularly evident in Basketball where players move between camera frustums and in Boxes where the moving object creates occlusions. This multi-view aggregation enables continuous tracking that would be impossible from any single viewpoint.}
    \label{fig:qualitative-results}
\end{figure*}
}

\newcommand{\lapaablation}{%
\begin{figure*}[t]
    \centering
    \includegraphics[width=\textwidth]{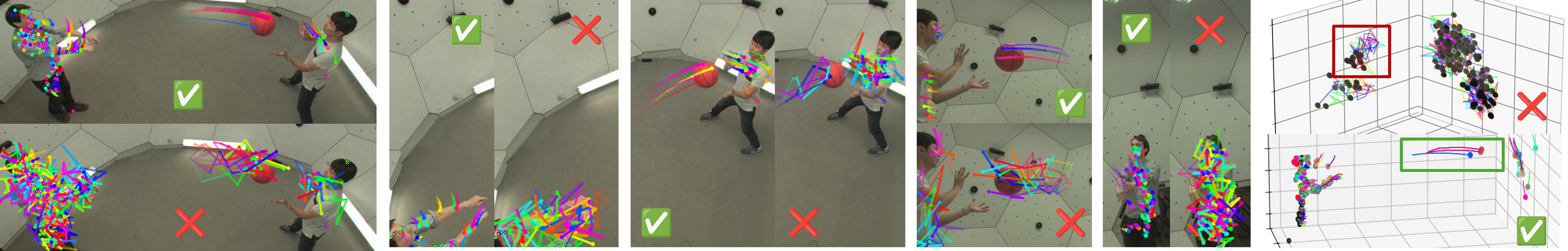}
    \caption{Ablation study demonstrating the impact of volumetric attention on temporal consistency. Each view is split into two parts: \greencheck{} shows results with attention mechanism enabled, while \redcross{} shows results with attention disabled. The model with attention maintains temporally coherent 3D track projections across all views, particularly evident in the mid-air ball sequence where the non-attention variant (\redcross{}) loses temporal coherence entirely, resulting in illogical track reconstructions. The attention mechanism (\greencheck{}) preserves consistent point identities and smooth trajectories even during challenging occlusions and rapid motion. This is especially pronounced in the 3D basketball trajectories marked with bounding boxes.}
    \label{fig:volumetric-attention}
\end{figure*}
}

\newcommand{\lapapipelinedetailed}{%
\begin{figure*}[t]
    \centering
    \includegraphics[width=\textwidth]{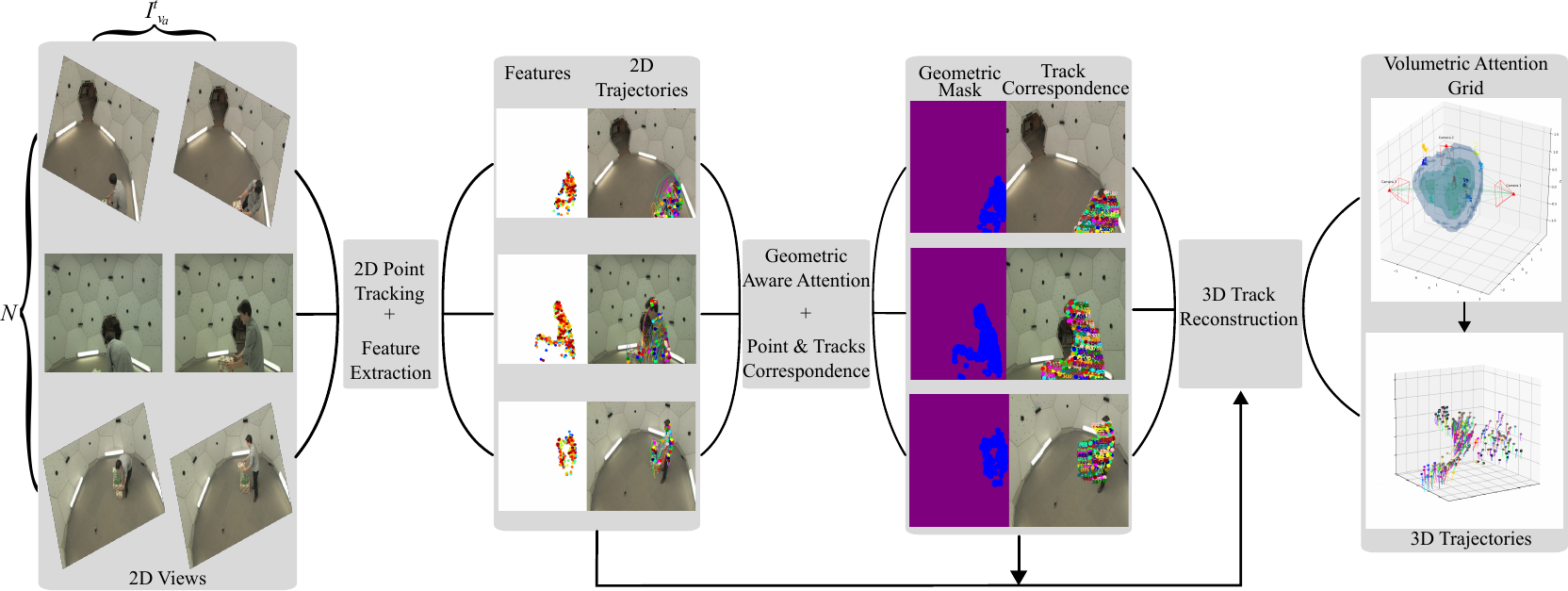}
    \caption{\textbf{Detailed LAPA Architecture}: Shows the complete data flow through our five-stage pipeline with volumetric attention grid visualization.}
    \label{fig:lapa-pipeline-detailed}
\end{figure*}
}

\newcommand{\camerasetupmultiview}{%
\begin{figure}[t]
    \centering
    \includegraphics[width=0.8\columnwidth]{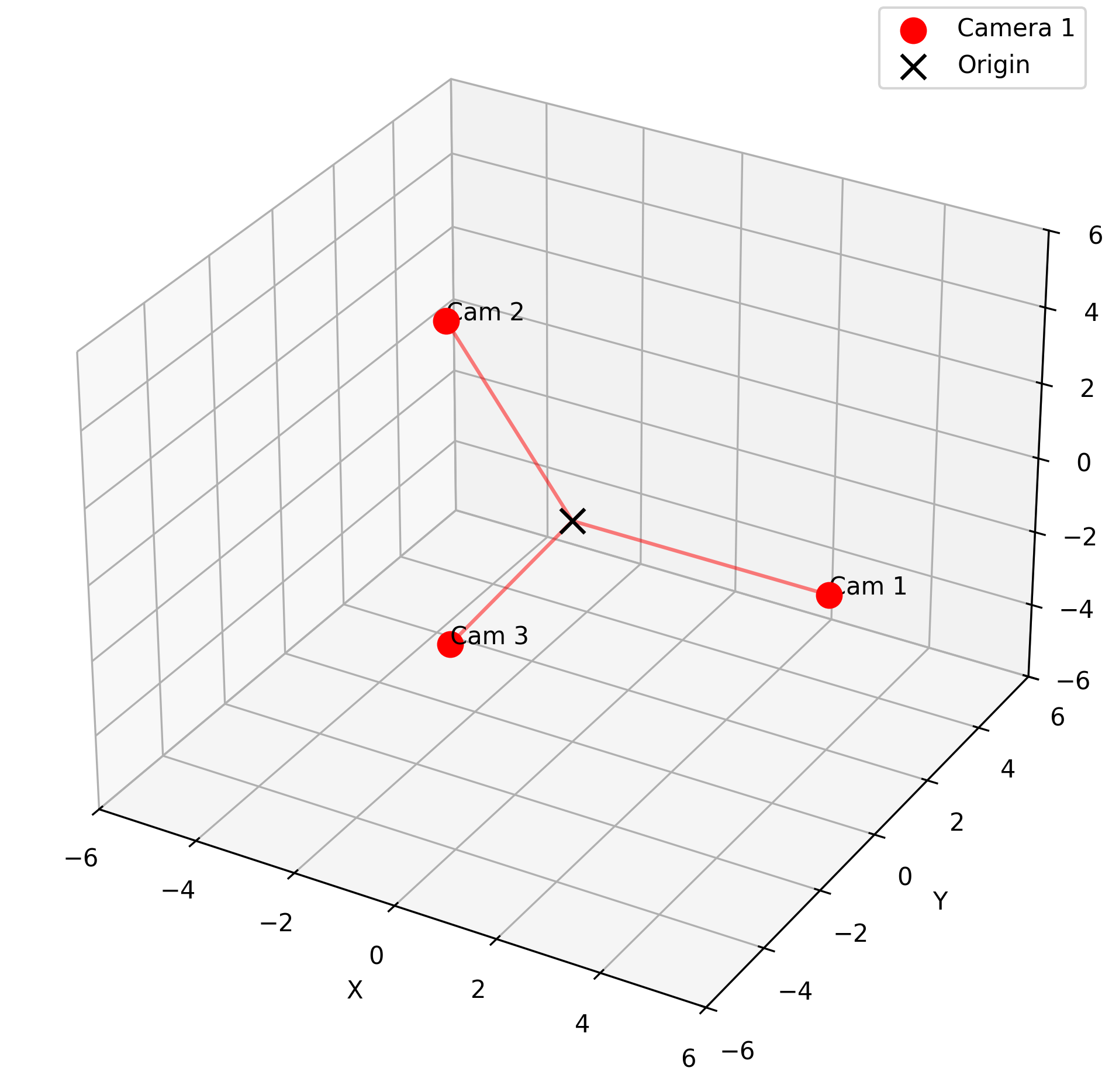}
    \caption{\textbf{Camera setup} for multi-view tracking with three cameras (red) positioned around the world origin (black X).}
    \label{fig:camera-setup}
\end{figure}
}
\newcommand{\comparison}{
\begin{table}[t]
\centering
\scriptsize 
\caption{Taxonomy of tracking methods revealing the research gap in multi-camera point tracking. LAPA uniquely fills this gap by combining point-level tracking with multi-camera support. \cmark: supported, \xmark: not supported, $\circ$: partially supported. Reconst.: 3D reconstruction, Bench.: introducing a benchmark dataset, Comp.: we provided quantitative comparison against these methods in our Experiments and Results section, Eval code: evaluation code publicly available.}
\label{tab:related_comparison}
\setlength{\tabcolsep}{2pt}
\renewcommand{\arraystretch}{1.2}
\begin{tabular}{l|cc|cccc|c}
\toprule
\multirow{2}{*}{\textbf{Method}} & \multicolumn{2}{c|}{\textbf{Tracking Type}} & \multicolumn{4}{c|}{\textbf{Capabilities}} & \textbf{Eval.} \\
& \textbf{Object} & \textbf{Point} & \textbf{Reconst.} & \textbf{Real-time} & \textbf{Bench.} & \textbf{Comp.} & \textbf{Code} \\
\midrule
\multicolumn{8}{c}{\cellcolor{blue!5}\textit{Single-Camera Trackers}} \\
CoTracker~\cite{karaev2024cotracker} & \xmark & \cmark & \xmark & \cmark & \xmark & \cmark & \cmark \\
Track-On~\cite{aydemir2025track} & \xmark & \cmark & \xmark & \cmark & \xmark & \cmark & \cmark \\
BootsTAPIR~\cite{doersch2024bootstap} & \xmark & \cmark & \xmark & \cmark & \xmark & \cmark & \cmark \\
SpatialTracker~\cite{xiao2024spatialtracker} & \xmark & \cmark & $\circ$ & \xmark & \xmark & \cmark & \cmark \\
SpatialTrackerV2~\cite{xiao2025spatialtrackerv2} & \xmark & \cmark & \cmark & \xmark & \xmark & \cmark & \cmark \\
TAPIP3D~\cite{zhang2025tapip3d} & \xmark & \cmark & \cmark & $\circ$ & \xmark & \cmark & \cmark \\
\midrule
\multicolumn{8}{c}{\cellcolor{green!5}\textit{Multi-Camera Trackers}} \\
MCTR~\cite{niculescu2025mctr} & \cmark & \xmark & \xmark & \cmark & \xmark & \xmark & \xmark \\
Graph-DETR4D~\cite{chen2024graph} & \cmark & \xmark & \cmark & $\circ$ & \xmark & \xmark & \cmark \\
\midrule
\multicolumn{8}{c}{\cellcolor{gray!15}\textit{\textbf{Research Gap: Multi-Camera Point Trackers}}} \\
\multicolumn{8}{c}{\includegraphics[height=1.5em]{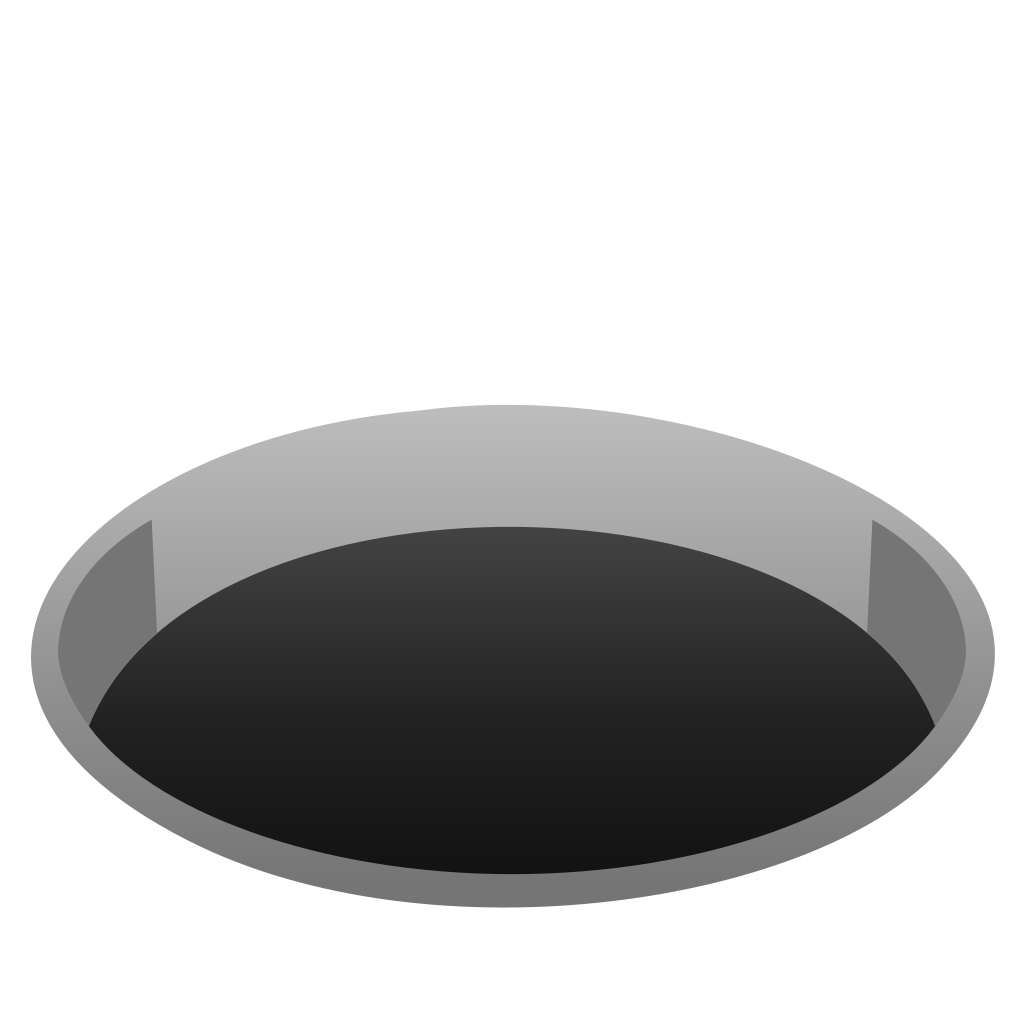}} \\
\midrule
\rowcolor{yellow!20}
\textbf{LAPA (Ours)} & \xmark & \cmark & \xmark & \cmark & \cmark & -- & \cmark \\
\bottomrule
\end{tabular}
\vspace{-0.2in}
\end{table}
}

\newcommand{\trackingMetricsTable}{
\begin{table*}[!htbp]
    \centering
    \small
    \caption{Quantitative evaluation of multi-camera point tracking performance on TAPVid-3D Panoptic and PointOdyssey datasets. We compare single-camera methods adapted for multi-camera setups through depth estimation or index matching against our unified LAPA approach. APD: Average Position Detection accuracy, OA: Occlusion Accuracy, 3D-AJ: 3D Average Jaccard, 2D-AJ: 2D Average Jaccard from projected tracks. Best results in \textbf{bold}. }
    \label{tab:tracking_results}
    \setlength{\tabcolsep}{2pt}
    \begin{tabular}{l|cccc|cccc}
    \toprule
    \multirow{2}{*}{\textbf{Method}} & \multicolumn{4}{c|}{\textbf{TAPVid-3D Panoptic \cite{koppula2024tapvid}}} & \multicolumn{4}{c}{\textbf{PointOdyssey \cite{pointodyssey2023}}} \\
     & APD$\uparrow$ & OA$\uparrow$ & 3D-AJ$\uparrow$ & 2D-AJ$\uparrow$ & APD$\uparrow$ & OA$\uparrow$ & 3D-AJ$\uparrow$ & 2D-AJ$\uparrow$ \\
    \midrule
    \multicolumn{9}{c}{\cellcolor{blue!5}\textit{Single-Camera Point Trackers with Depth Estimation}} \\
    CoTracker \cite{karaev2024cotracker} + ZoeDepth & 14.8 & 83.4 & 8.7 & 19.4 & 84.3 & 80.8 & 0.76 & 15.9 \\
    CoTracker \cite{karaev2024cotracker} + COLMAP & 15.0 & 79.1 & 9.3 & 13.5 & 82.1 & 79.2 & 0.74 & 14.1 \\
    TrackOn \cite{aydemir2025track} + ZoeDepth & 13.8 & 80.0 & 8.3 & 18.2 & 85.7 & 82.4 & 0.78 & 16.4 \\
    TrackOn \cite{aydemir2025track} + COLMAP & 12.4 & 76.1 & 7.4 & 10.7 & 86.9 & 83.7 & 0.81 & 13.8 \\
    BootsTAPIR \cite{doersch2024bootstap} + ZoeDepth & 14.8 & 84.8 & 8.8 & 19.0 & 83.6 & 81.1 & 0.75 & 15.7 \\
    BootsTAPIR \cite{doersch2024bootstap} + COLMAP & 14.9 & 81.4 & 9.3 & 11.6 & 84.8 & 82.3 & 0.77 & 14.3 \\
    \midrule
    \multicolumn{9}{c}{\cellcolor{green!5}\textit{Single-Camera Point Trackerswith 3D Estimation/Reconstruction}} \\
    SpatialTracker \cite{xiao2024spatialtracker} & 15.5 & 83.7 & 9.0 & 19.2 & 82.1 & 79.4 & 0.73 & 15.5 \\
    SpatialTrackerV2 \cite{xiao2025spatialtrackerv2} & 28.7 & 86.1 & 18.6 & 28.7 & 87.4 & 85.2 & 0.89 & 17.1 \\
    TAPIP3D \cite{zhang2025tapip3d} & 32.1 & 87.4 & 21.9 & 32.1 & 88.1 & 86.3 & 0.92 & 17.8 \\
    \midrule
    \rowcolor[gray]{0.9} \textbf{LAPA (Ours)} & \textbf{37.5} & \textbf{90.3} & \textbf{30.2} & \textbf{40.1} & \textbf{90.3} & \textbf{88.7} & \textbf{0.95} & \textbf{18.5} \\
    \bottomrule
    \end{tabular}
    \vspace{-0.1in}
\end{table*}
}

\newcommand{\ablationTable}{
\begin{table}[t]
\centering
\small
\caption{Ablation study on LAPA components. We incrementally integrate components into the baseline model to train and evaluate the model's performance. Best results in \textbf{bold}.}
\label{tab:ablation}
\setlength{\tabcolsep}{3pt}
\renewcommand{\arraystretch}{1.2}
\begin{tabular}{l|cccc}
\toprule
\textbf{Configuration} & \textbf{APD}$\uparrow$ & \textbf{OA}$\uparrow$ & \textbf{3D-AJ}$\uparrow$ & \textbf{2D-AJ}$\uparrow$ \\
\midrule
Baseline (B) & 8.2 & 58.3 & 5.8 & 12.7 \\
B + Attention (A) & 12.4 & 67.2 & 6.5 & 15.3 \\
BA + Geometry (G) & 18.1 & 74.8 & 7.3 & 18.5 \\
BAG + 2-Cam & 24.2 & 79.1 & 7.6 & 21.8 \\
\rowcolor{gray!10} \textbf{Full LAPA (3-cam)} & \textbf{37.5} & \textbf{90.3} & \textbf{30.2} & \textbf{40.1} \\
\bottomrule
\end{tabular}
\end{table}
}

\newcommand{\cameraAblationTable}{
\begin{table}[t]
\centering
\small
\caption{Impact of camera count $C$ on tracking performance. Results show performance peaks at 3 cameras with diminishing returns beyond $C>3$. Best results in \textbf{bold}.}
\label{tab:camera_ablation}
\setlength{\tabcolsep}{3pt}
\renewcommand{\arraystretch}{1.2}
\begin{tabular}{l|cccc}
\toprule
\textbf{$C$} & \textbf{APD}$\uparrow$ & \textbf{OA}$\uparrow$ & \textbf{3D-AJ}$\uparrow$ & \textbf{2D-AJ}$\uparrow$ \\
\midrule
2 & 24.2 & 79.1 & 7.6 & 21.8 \\
\rowcolor{gray!10} \textbf{3} & \textbf{37.5} & \textbf{90.3} & \textbf{30.2} & \textbf{40.1} \\
4 & 36.8 & 84.5 & 29.1 & 35.7 \\
5 & 35.9 & 83.2 & 28.5 & 34.1 \\
\bottomrule
\end{tabular}
\end{table}
}

\newcommand{\attentionGeometryAblationTables}{
\begin{table}[t]
    \centering
    \small
    \caption{Ablation Studies on Attention and Geometry. Volumetric attention and SfM constraints significantly enhance performance. Best results are in \textbf{bold}.}
    \label{tab:combined_ablation}
    \setlength{\tabcolsep}{3pt}
    \begin{tabular}{l|cccc}
    \toprule
    \textbf{Components} & \textbf{APD}$\uparrow$ & \textbf{OA}$\uparrow$ & \textbf{3D-AJ}$\uparrow$ & \textbf{2D-AJ}$\uparrow$ \\
    \midrule
    \multicolumn{5}{c}{\cellcolor{blue!5}\textit{Effectiveness of Volumetric Attention (VA)}} \\
    Without VA & 15.9 & 67.9 & 6.8 & 18.7 \\
   \rowcolor{gray!10} \textbf{With VA} & \textbf{37.5} & \textbf{90.3} & \textbf{30.2} & \textbf{40.1} \\
    \midrule
    \multicolumn{5}{c}{\cellcolor{green!5}\textit{Comparison of Geometric Constraints}} \\
    Epipolar & 22.3 & 79.4 & 7.6 & 24.5 \\
   \rowcolor{gray!10} \textbf{SfM} & \textbf{37.5} & \textbf{90.3} & \textbf{30.2} & \textbf{40.1} \\
    \bottomrule
    \end{tabular}
    \vspace{-0.1in}
\end{table}
}

\newcommand{\gridAblationTable}{
\begin{table}[t]
\centering
\small
\caption{Effect of volumetric grid resolution. A $16 \times 16 \times 16$ grid provides the best balance between performance and computational efficiency. Best results in \textbf{bold}.}
\label{tab:grid_ablation}
\setlength{\tabcolsep}{3pt}
\renewcommand{\arraystretch}{1.2}
\begin{tabular}{l|cccc}
\toprule
\textbf{Grid Size} & \textbf{APD}$\uparrow$ & \textbf{OA}$\uparrow$ & \textbf{3D-AJ}$\uparrow$ & \textbf{2D-AJ}$\uparrow$ \\
\midrule
$8 \times 8 \times 8$ & 32.4 & 76.3 & 27.5 & 32.8 \\
\rowcolor{gray!10} $\textbf{16} \times \textbf{16} \times \textbf{16}$ & \textbf{37.5} & \textbf{90.3} & \textbf{30.2} & \textbf{40.1} \\
$24 \times 24 \times 24$ & 36.8 & 85.9 & 29.8 & 36.1 \\
\bottomrule
\end{tabular}
\end{table}
}

\newcommand{\lossComponentsAblationTable}{
\begin{table}[t]
\centering
\small
\caption{Contribution of different combinations of loss components achieved by setting their respective $\lambda=0$. Best results in \textbf{bold}.}
\label{tab:loss_ablation}
\setlength{\tabcolsep}{3pt}
\renewcommand{\arraystretch}{1.2}
\begin{tabular}{ccc|cccc}
\toprule
\multicolumn{3}{c|}{\textbf{Components}} & \multirow{2}{*}{\textbf{APD}$\uparrow$} & \multirow{2}{*}{\textbf{OA}$\uparrow$} & \multirow{2}{*}{\textbf{3D-AJ}$\uparrow$} & \multirow{2}{*}{\textbf{2D-AJ}$\uparrow$} \\
Recon & Proj & Attn &  &  &  &  \\
\midrule
\textcolor{gray}{\ding{55}} & \checkmark & \checkmark & 18.1 & 62.3 & 6.1 & 19.5 \\
\checkmark & \textcolor{gray}{\ding{55}} & \checkmark & 28.3 & 78.6 & 7.5 & 28.9 \\
\checkmark & \checkmark & \textcolor{gray}{\ding{55}} & 31.1 & 80.9 & 7.8 & 32.8 \\
\rowcolor{gray!10} \checkmark & \checkmark & \checkmark & \textbf{37.5} & \textbf{90.3} & \textbf{30.2} & \textbf{40.1} \\
\bottomrule
\end{tabular}
\end{table}
}
\newcommand{\optimizedLossTable}{
\begin{table*}[t]
\centering
\small
\caption{Optimized loss weights for LAPA with 3-camera configuration. Best results in \textbf{bold}.}
\label{tab:optimized_loss}
\setlength{\tabcolsep}{3pt}
\renewcommand{\arraystretch}{1.2}
\begin{tabular}{l|ccc|cccc}
\toprule
\multirow{2}{*}{\textbf{Configuration}} & \multicolumn{3}{c|}{\textbf{Loss Weights}} & \multirow{2}{*}{\textbf{APD}$\uparrow$} & \multirow{2}{*}{\textbf{OA}$\uparrow$} & \multirow{2}{*}{\textbf{3D-AJ}$\uparrow$} & \multirow{2}{*}{\textbf{2D-AJ}$\uparrow$} \\
 & $\lambda_\text{recon}$ & $\lambda_\text{proj}$ & $\lambda_\text{attn}$ &  &  &  &  \\
\midrule
Default & 1.0 & 0.5 & 0.5 & 37.5 & 86.3 & 30.2 & 37.4 \\
Balanced & 1.0 & 1.0 & 1.0 & 36.7 & 85.4 & 29.1 & 35.8 \\
Recon-Heavy & 2.0 & 0.5 & 0.5 & 37.8 & 86.5 & 30.8 & 38.2 \\
Attn-Heavy & 1.0 & 0.5 & 1.0 & 38.1 & 86.8 & 31.1 & 38.6 \\
Proj-Heavy & 1.0 & 1.0 & 0.5 & 36.3 & 85.9 & 28.9 & 36.3 \\
Proj+Attn & 1.0 & 1.0 & 1.0 & 37.6 & 86.1 & 30.3 & 37.5 \\
\rowcolor{gray!10} Optimized & 1.0 & 0.7 & 0.8 & \textbf{39.1} & \textbf{87.2} & \textbf{32.6} & \textbf{40.1} \\
\bottomrule
\end{tabular}
\end{table*}
}

\newcommand{\comprehensiveAblationTable}{
\begin{table*}[t]
\centering
\small
\caption{Comprehensive Ablation Study of LAPA Architecture: This table shows the impact of each component across all evaluation metrics. Best results in each category in \textbf{bold}.}
\label{tab:S_comprehensive_ablation}
\setlength{\tabcolsep}{3pt}
\renewcommand{\arraystretch}{1.2}
\begin{tabular}{l|l|cccc}
\toprule
\textbf{Component} & \textbf{Setting} & \textbf{APD}$\uparrow$ & \textbf{OA}$\uparrow$ & \textbf{3D-AJ}$\uparrow$ & \textbf{2D-AJ}$\uparrow$ \\
\midrule
\multirow{2}{*}{Attention} & Without & 15.9 & 67.9 & 6.8 & 18.7 \\
 & \cellcolor{gray!10}\textbf{With} & \cellcolor{gray!10}\textbf{37.5} & \cellcolor{gray!10}\textbf{90.3} & \cellcolor{gray!10}\textbf{30.2} & \cellcolor{gray!10}\textbf{40.1} \\
\midrule
\multirow{2}{*}{Geometry} & Epipolar & 22.3 & 79.4 & 7.6 & 24.5 \\
 & \cellcolor{gray!10}\textbf{SfM} & \cellcolor{gray!10}\textbf{37.5} & \cellcolor{gray!10}\textbf{90.3} & \cellcolor{gray!10}\textbf{30.2} & \cellcolor{gray!10}\textbf{40.1} \\
\midrule
\multirow{4}{*}{Cameras} & 2 & 24.2 & 65.8 & 7.6 & 21.8 \\
 & \cellcolor{gray!10}\textbf{3} & \cellcolor{gray!10}\textbf{37.5} & \cellcolor{gray!10}\textbf{90.3} & \cellcolor{gray!10}\textbf{30.2} & \cellcolor{gray!10}\textbf{40.1} \\
 & 4 & 36.8 & 84.5 & 29.1 & 35.7 \\
 & 5 & 35.9 & 83.2 & 28.5 & 34.1 \\
\midrule
\multirow{3}{*}{Grid Size} & $8 \times 8 \times 8$ & 32.4 & 76.3 & 27.5 & 32.8 \\
 & \cellcolor{gray!10}$\textbf{16} \times \textbf{16} \times \textbf{16}$ & \cellcolor{gray!10}\textbf{37.5} & \cellcolor{gray!10}\textbf{86.3} & \cellcolor{gray!10}\textbf{30.2} & \cellcolor{gray!10}\textbf{37.4} \\
 & $24 \times 24 \times 24$ & 36.8 & 85.9 & 29.8 & 36.1 \\
\midrule
\multirow{7}{*}{Loss Weights} & Default (1.0, 0.5, 0.5) & 37.5 & 86.3 & 30.2 & 37.4 \\
 & Balanced (1.0, 1.0, 1.0) & 36.7 & 85.4 & 29.1 & 35.8 \\
 & Recon-Heavy (2.0, 0.5, 0.5) & 37.8 & 86.5 & 30.8 & 38.2 \\
 & Attn-Heavy (1.0, 0.5, 1.0) & 38.1 & 86.8 & 31.1 & 38.6 \\
 & Proj-Heavy (1.0, 1.0, 0.5) & 36.3 & 85.9 & 28.9 & 36.3 \\
 & Proj+Attn (1.0, 1.0, 1.0) & 37.6 & 86.1 & 30.3 & 37.5 \\
 & \cellcolor{gray!10}\textbf{Optimized (1.0, 0.7, 0.8)} & \cellcolor{gray!10}\textbf{39.1} & \cellcolor{gray!10}\textbf{87.2} & \cellcolor{gray!10}\textbf{32.6} & \cellcolor{gray!10}\textbf{40.1} \\
\bottomrule
\end{tabular}
\end{table*}
}

\newcommand{\performancetable}{
\begin{table*}[t]
\centering
\small
\caption{Computational performance analysis of LAPA architecture across different configurations and hardware platforms on the TAPVid-3D CMU Panoptic dataset. $\dagger$ indicates the recommended configuration for optimal performance-accuracy trade-off. Depth estimation methods (+ZoeDepth, +COLMAP) significantly increase computational overhead. }
\label{tab:S_computational_performance}
\setlength{\tabcolsep}{2.5pt}
\renewcommand{\arraystretch}{1.2}
\begin{tabular}{lcccccc|c}
\toprule
\multirow{2}{*}{\textbf{Configuration}} & \multirow{2}{*}{\textbf{APD}$\uparrow$} & \multirow{2}{*}{\textbf{Params}} & \multirow{2}{*}{\textbf{FLOPS}$\downarrow$} & \multicolumn{2}{c}{\textbf{FPS}$\uparrow$} & \multirow{2}{*}{\textbf{Memory}$\downarrow$} & \multirow{2}{*}{\textbf{Attention}} \\
\cmidrule(lr){5-6}
 & & & & \textbf{RTX Titan} & \textbf{V100} & & \\
\midrule
\multicolumn{8}{l}{\cellcolor{blue!5}\textit{LAPA - Grid Resolution: $16 \times 16 \times 16$}} \\
LAPA (2 cameras) & 24.2\% & 372.7K & 28.3M & 76.3 & 105.2 & 1.8GB & \checkmark \\
LAPA (3 cameras)$\dagger$ & 37.5\% & 372.7K & 42.5M & 39.0 & 55.4 & 2.6GB & \checkmark \\
LAPA (4 cameras) & 36.8\% & 372.7K & 56.7M & 26.0 & 37.9 & 3.4GB & \checkmark \\
LAPA (5 cameras) & 35.9\% & 372.7K & 70.8M & 17.8 & 26.3 & 4.2GB & \checkmark \\
\midrule
\multicolumn{8}{l}{\cellcolor{green!5}\textit{LAPA - Grid Resolution: $24 \times 24 \times 24$}} \\
LAPA (2 cameras) & 24.0\% & 372.7K & 95.6M & 32.4 & 45.8 & 3.6GB & \checkmark \\
LAPA (3 cameras) & 37.2\% & 372.7K & 143.3M & 21.7 & 31.2 & 5.2GB & \checkmark \\
LAPA (4 cameras) & 36.5\% & 372.7K & 191.1M & 15.8 & 22.9 & 6.8GB & \checkmark \\
LAPA (5 cameras) & 35.7\% & 372.7K & 238.9M & 12.1 & 17.6 & 8.5GB & \checkmark \\
\midrule
\multicolumn{8}{l}{\cellcolor{yellow!5}\textit{LAPA Without Attention ($16 \times 16 \times 16$)}} \\
LAPA-Base (2 cameras) & 15.2\% & 41.5K & 13.5M & 85.3 & 123.4 & 0.8GB & \textcolor{gray}{\ding{55}} \\
LAPA-Base (3 cameras) & 18.7\% & 41.5K & 20.2M & 63.1 & 91.7 & 1.1GB & \textcolor{gray}{\ding{55}} \\
LAPA-Base (4 cameras) & 22.1\% & 41.5K & 27.0M & 49.3 & 72.6 & 1.4GB & \textcolor{gray}{\ding{55}} \\
LAPA-Base (5 cameras) & 22.3\% & 41.5K & 33.7M & 39.2 & 58.9 & 1.7GB & \textcolor{gray}{\ding{55}} \\
\midrule
\multicolumn{8}{l}{\cellcolor{red!5}\textit{Single-Camera Methods with Depth Estimation (Much Slower)}} \\
CoTracker + ZoeDepth & 14.8\% & 245K & 89.2M & 8.7 & 12.3 & 4.1GB & \textcolor{gray}{\ding{55}} \\
CoTracker + COLMAP & 15.0\% & 198K & 76.5M & 11.2 & 15.8 & 3.8GB & \textcolor{gray}{\ding{55}} \\
TrackOn + ZoeDepth & 13.8\% & 287K & 94.1M & 7.9 & 11.1 & 4.3GB & \textcolor{gray}{\ding{55}} \\
TrackOn + COLMAP & 12.4\% & 231K & 81.7M & 9.8 & 13.7 & 4.0GB & \textcolor{gray}{\ding{55}} \\
BootsTAPIR + ZoeDepth & 14.8\% & 269K & 91.8M & 8.2 & 11.6 & 4.2GB & \textcolor{gray}{\ding{55}} \\
BootsTAPIR + COLMAP & 14.9\% & 215K & 78.9M & 10.5 & 14.9 & 3.9GB & \textcolor{gray}{\ding{55}} \\
\midrule
\multicolumn{8}{l}{\textit{Baseline Methods - Single-Camera (Multi-Camera Setup)}} \\
CoTracker \cite{karaev2024cotracker} & 15.0\% & 78K & 12.0M & 95.4 & 137.8 & 1.3GB & \textcolor{gray}{\ding{55}} \\
TrackOn \cite{aydemir2025track} & 12.4\% & 150K & 28.0M & 41.2 & 59.5 & 2.2GB & \checkmark \\
\midrule
\multicolumn{8}{l}{\textit{Baseline Methods - Single-Camera with 3D}} \\
SpatialTracker \cite{xiao2024spatialtracker} & 15.5\% & 95K & 18.5M & 52.3 & 73.8 & 1.9GB & \textcolor{gray}{\ding{55}} \\
SpatialTrackerV2 \cite{xiao2025spatialtrackerv2} & 28.7\% & 124K & 45.2M & 3.2 & 4.8 & 2.8GB & \textcolor{gray}{\ding{55}} \\
TAPIP3D \cite{zhang2025tapip3d} & 32.1\% & 187K & 52.7M & 2.9 & 4.1 & 3.2GB & \textcolor{gray}{\ding{55}} \\
\bottomrule
\end{tabular}
\end{table*}
}

\newcommand{\calibrationrobustnesstable}{
\begin{table}[h]
\centering
\small
\caption{Performance under calibration noise (3-camera setup on TAPVid-3D-MC).}
\label{tab:S_calibration_robustness}
\begin{tabular}{l|c|cc}
\hline
\textbf{Noise Type} & \textbf{Level} & \textbf{APD↑} & \textbf{3D-AJ↑} \\
\hline
Perfect Calibration & - & 37.5 & 30.2 \\
\hline
Intrinsic & $\sigma = 1.0$ px & 32.1 & 26.1 \\
Rotation & $\sigma = 1.0$° & 33.7 & 27.3 \\
Translation & $\sigma = 5$ cm & 31.8 & 24.9 \\
\hline
Combined & Typical noise & 28.4 & 22.7 \\
\hline
\end{tabular}
\end{table}
}
\usepackage{url}
\usepackage{booktabs} 
\usepackage{amsfonts}
\usepackage{nicefrac}
\usepackage{microtype}
\usepackage{placeins}
\usepackage{pifont}
\usepackage{float}
\newcommand{\cmark}{\ding{51}}
\newcommand{\xmark}{\ding{55}}
\usepackage{multirow}
\usepackage{colortbl}
\usepackage{xcolor}
\usepackage{makecell}

\usepackage{subcaption}
\usepackage{graphicx}
\usepackage{fontawesome5}
\newcommand{\redcross}{\includegraphics[height=1em]{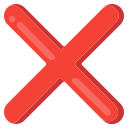}}
\newcommand{\greencheck}{\includegraphics[height=1em]{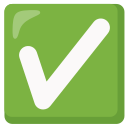}}
\definecolor{cvprblue}{rgb}{0.21,0.49,0.74}
\usepackage[pagebackref,breaklinks,colorlinks,citecolor=cvprblue]{hyperref}
\def\paperID{105} % *** Enter the Paper ID here
\def\confName{3DV\xspace}
\def\confYear{2026\xspace}
\title{Look Around and Pay Attention: \\ Multi-camera Point Tracking Reimagined with Transformers}
\author{Bishoy Galoaa, Xiangyu Bai, Shayda Moezzi, Utsav Nandi, \\
Sai Siddhartha Vivek Dhir Rangoju, Somaieh Amraee, Sarah Ostadabbas\\
Northeastern University\\
Boston, MA, USA\\
{\tt\small \{galoaa.b, bai.xiang, moezzi.s, nandi.u, rangoju.s, amraee.s, s.ostadabbas\}@northeastern.edu}
}
\makeatletter
\def\@maketitle{
   \newpage
   \null
   \iftoggle{cvprrebuttal}{\vspace*{-.3in}}{\vskip .375in}
   \begin{center}
      \iftoggle{cvprrebuttal}{{\large \bf \@title \par}}{{\Large \bf \@title \par}}
      \iftoggle{cvprrebuttal}{\vspace*{-22pt}}{\vspace*{24pt}}{
        \large
        \lineskip .5em
        \begin{tabular}[t]{c}
          \iftoggle{cvprfinal}{
            \@author
          }{
            \iftoggle{cvprrebuttal}{}{
              Anonymous \confName~submission\\
              \vspace*{1pt}\\
              Paper ID \paperID
            }
          }
        \end{tabular}
        \par
      }
      \vskip .5em
      \vspace*{12pt}
   \end{center}
   \vspace{0.5cm}
   \begin{center}
      \includegraphics[width=\textwidth]{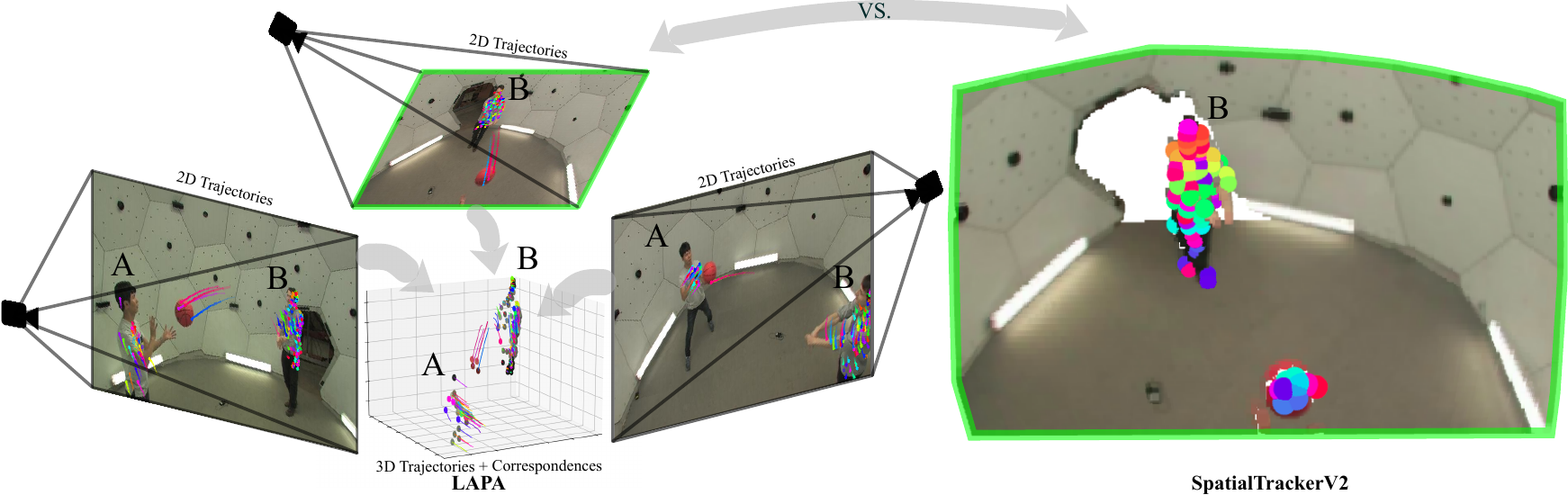}
      \captionof{figure}{\textbf{LAPA vs. SpatialTrackerV2}: Given the same scene, single-view methods fail completely when subject (A) is outside their camera's field of view, while LAPA maintains tracking using multiple cameras. Left: LAPA corresponds 2D trajectories from all views into consistent 3D trajectories. Right: SpatialTrackerV2 \cite{xiao2025spatialtrackerv2} depends on its camera's limited field of view, unable to track or reconstruct subjects beyond its visible range. Green boxes indicate that the scene is taken from same camera perspective in LAPA and SpatialTrackerV2.}
      \label{fig:lapa-concept}
   \end{center}
   \vspace{0.5cm}
}

\makeatother
\begin{document}
\maketitle
\begin{abstract}
This paper presents LAPA (Look Around and Pay Attention), a novel end-to-end transformer-based architecture for multi-camera point tracking that integrates appearance-based matching with geometric constraints. Traditional pipelines decouple detection, association, and tracking, leading to error propagation and temporal inconsistency in challenging scenarios. LAPA addresses these limitations by leveraging attention mechanisms to jointly reason across views and time, establishing soft correspondences through a cross-view attention mechanism enhanced with geometric priors. Instead of relying on classical triangulation, we construct 3D point representations via attention-weighted aggregation, inherently accommodating uncertainty and partial observations. Temporal consistency is further maintained through a transformer decoder that models long-range dependencies, preserving identities through extended occlusions. Extensive experiments on challenging datasets, including our newly created multi-camera (MC) versions of TAPVid-3D panoptic and PointOdyssey, demonstrate that our unified approach significantly outperforms existing methods, achieving 37.5\% APD on TAPVid-3D-MC and 90.3\% APD on PointOdyssey-MC, particularly excelling in scenarios with complex motions and occlusions. Code is available at \url{https://github.com/ostadabbas/Look-Around-and-Pay-Attention-LAPA-}.
\end{abstract}    
\section{Introduction}
Tracking points through video has become a universal scene representation in robotics \cite{lou2024robogs, niu2025pretraining, tang2024kalib}, reconstruction \cite{kasten2024fast, lei2024mosca}, and augmented reality \cite{klein2007parallel, li2017monocular}. While transformer-based methods achieve impressive accuracy in single-view settings \cite{karaev2024cotracker, aydemir2025track}, they fail when tracked points become occluded \cite{tumanyan2024dino}. A natural solution is to use multiple cameras. Yet, despite the ubiquity of multi-camera deployments, no existing method performs end-to-end point tracking across multiple calibrated cameras.

This gap is surprising given the clear advantages of multi-camera systems. Consider Figure~\ref{fig:lapa-concept}, where a person walks through a multi-camera setup: Single-view reconstruction-based methods such as SpatialTrackerV2 \cite{xiao2025spatialtrackerv2} fail completely when subject (A) leaves the camera's field of view. Even with 3D reconstruction capabilities, they cannot track points that are not visible. For example the basketball appears visible in one view, but the recipient (Subject A) is entirely outside the camera's perspective. Continuous 3D tracking is critical in applications such as robotic manipulation \cite{sekkat2021vision}, where contact points must be tracked precisely; surveillance, where facial keypoints need to be followed across cameras\cite{zhu2024camera}; and sports analytics \cite{bridgeman2019multi}, requiring accurate joint trajectories during complex motion. These applications demand more than bounding boxes--they require precise, temporally consistent tracking of specific points across all available views. In contrast, LAPA leverages all views simultaneously: when one camera loses sight of the subject, other views maintain visibility, enabling continuous 3D tracking.

The lack of multi-camera point tracking solutions stems from several technical challenges. First, cross-view correspondence requires reasoning about 3D geometry while accommodating appearance changes and partial occlusions. Second, temporal consistency becomes complex when points appear and disappear across views at different times. Third, traditional sequential pipelines--detecting points, establishing correspondences, then triangulating--are prone to cumulative error. Recent works have attempted to address some of these challenges in isolation. SpatialTracker \cite{xiao2024spatialtracker} and TAPIP3D \cite{zhang2025tapip3d} lift 2D tracks to 3D using monocular depth or persistent 3D representations, but remain limited by single-view inputs. Multi-view correspondence methods such as EpiTransformer \cite{he2020epipolar} and MVSTER \cite{wang2022mvster} find matches across cameras but do not perform temporal tracking. Multi-camera object trackers \cite{niculescu2025mctr, zhang2022multicamera} maintain identities across views but operate at the bounding box level, lacking the fine-grained detail required for precise applications.

We present LAPA (Look Around and Pay Attention), the first end-to-end transformer-based architecture for multi-camera point tracking. As illustrated in Figure~\ref{fig:lapa-concept}, LAPA takes synchronized multi-view video as input and directly outputs consistent 3D point trajectories, avoiding the error propagation of sequential approaches. The system fuses 2D trajectories from all views into complete 3D tracks, preserving correspondences even when points become occluded in individual views. Our core insight is that spatial proximity in 3D space is often a more reliable cue for correspondence than appearance matching, particularly under severe occlusion or viewpoint changes. 

To this end, LAPA introduces a distance-based volumetric attention mechanism that incorporates epipolar geometry while remaining fully differentiable. Unlike traditional methods that make hard correspondence decisions, our approach uses soft attention weights to represent uncertainty, allowing robust handling of ambiguous matches. The volumetric representation enables direct reasoning of 3D spatial relationships, and cross-view attention ensures that all cameras inform tracking decisions. When a point disappears from one view, attention shifts to views where it remains visible, maintaining continuous tracking through occlusion.

Our work establishes multi-camera point tracking as a distinct computer vision task, separate from both multi-object tracking (bounding box level) and single-view point tracking (no geometric constraints). This opens new opportunities for applications requiring precise spatial understanding from multiple viewpoints, from detailed motion capture without markers to robust sensing in cluttered environments.

\noindent\textbf{Our contributions can be summarized as following:}
\begin{itemize}
\item We identify and formally define multi-camera point tracking as a critical gap in computer vision, distinct from object-level or single-view approaches, and show why existing methods cannot address this need.

\item We introduce LAPA, a novel end-to-end framework with distance-based volumetric attention that jointly optimizes detection, correspondence, and tracking across multiple cameras, eliminating cascading errors of traditional pipelines while achieving real-time performance (up to 39 FPS, depending on the number of cameras).

\item We develop procedural protocols to extend TAPVid-3D \cite{koppula2024tapvid} and PointOdyssey \cite{pointodyssey2023} into multi-camera benchmarks (TAPVid-3D-MC and PointOdyssey-MC), creating the first evaluation framework for multi-camera point tracking.

\item We achieve substantial improvements over adapted baseline models, with LAPA reaching 37.5\% APD on TAPVid-3D-MC and 90.3\% APD on PointOdyssey-MC, with particularly strong gains in challenging occlusion scenarios where traditional single-view methods fail.
\end{itemize}

\section{Related Work}

We position LAPA at the intersection of point tracking and multi-view geometry, establishing multi-camera point tracking as a distinct problem that demands specialized solutions. We review three categories of prior work to highlight the gap LAPA fills: single-camera point tracking (precise but unable to handle occlusions), multi-camera object tracking (occlusion-robust but lacking point-level precision), and multi-view correspondence (spatial matching without temporal tracking). 

\noindent
\textbf{Single-Camera Point Tracking.} Recent methods such as CoTracker~\cite{karaev2024cotracker} jointly track multiple points using shared context, improving robustness compared to tracking points independently. Track-On~\cite{aydemir2025track} extends this with transformer-based memory mechanisms for long-term tracking,while BootsTAPIR~\cite{doersch2024bootstap} provides robust tracking through bootstrap training techniques. DINO-Tracker~\cite{tumanyan2024dino} leverages pre-trained vision models for strong feature matching. To incorporate 3D understanding, SpatialTracker~\cite{xiao2024spatialtracker} and its successor SpatialTrackerV2~\cite{xiao2025spatialtrackerv2} attempt 3D reasoning from single views through depth estimation, with V2 achieving high accuracy but requiring 10-20 seconds per sequence. TAPIP3D~\cite{zhang2025tapip3d} maintains persistent 3D geometry representations to improve occlusion handling. However, all single-view methods fundamentally fail when a point is completely occluded--they must either guess or stop tracking. In contrast, multi-camera systems inherently solve this problem through complementary viewpoints, maintaining visibility when individual cameras lose sight of points.

\noindent
\textbf{Multi-Camera Object Tracking.} Methods like MCTR~\cite{niculescu2025mctr} track whole objects across cameras using bounding boxes and transformer-based appearance matching. Graph-DETR4D~\cite{chen2024graph} extends this approach to 3D object detection with spatio-temporal graph modeling, although its focus on autonomous driving datasets limits direct comparison. Graph-based methods~\cite{zhang2022multicamera} propagate information across camera networks, while probabilistic methods model uncertainty in cross-camera associations \cite{niculescu2025mctr, zhou2024ua}. The key limitation here is granularity: bounding boxes cannot provide the point-level precision required for robotic manipulation~\cite{sekkat2021vision} or detailed motion analysis \cite{stanczyk2025no}. Tracking a person across cameras is fundamentally different from tracking the exact position of their fingertips or tool contact points. 

\noindent
\textbf{Multi-View Correspondence.} Multi-view correspondence techniques focus on pixel-level matching across camera views, supplying critical geometric constraints relevant to our work. EpiTransformer~\cite{he2020epipolar} integrates epipolar geometry into attention mechanisms for robust feature matching, while MVSTER~\cite{wang2022mvster} extends this to dense multi-view stereo reconstruction using cross-attention along epipolar lines. Classical methods rely on fundamental matrices~\cite{hartley2003multiple} and feature descriptors~\cite{lowe2004distinctive} for establishing correspondences. These approaches excel at spatial correspondence but operate on individual frames, lacking temporal coherence--answering "which pixels match across views?" without addressing "how did this point move over time?" As a result, they are unsuitable for tracking tasks that require persistent point identities across frames. Nevertheless, their geometric principles, especially epipolar constraints, inform our attention mechanism design.

\comparison
As shown in Table~\ref{tab:related_comparison}, LAPA uniquely addresses the research gap in multi-camera point tracking. Single-camera trackers (CoTracker, Track-On, SpatialTracker, etc.) excel at point-level tracking but cannot leverage multiple viewpoints. Multi-camera trackers (MCTR, Graph-DETR4D) handle multiple cameras but only track whole objects. LAPA is the first to combine fine-grained point precision with full multi-camera capability. The absence of prior work in this space is underscored by the lack of dedicated benchmarks--existing datasets target either single-view point tracking or multi-camera object detection--motivating our creation of TAPVid-3D-MC and PointOdyssey-MC to aid research in this underexplored domain.
\lapapipeline

\section{Introducing LAPA}
\textbf{Problem Formulation} Given synchronized multi-view frames $\{I_{v_a}\}_{a=1}^N$ from $N$ calibrated cameras, where $v_a$ indexes view $a \in \{1,2,...,N\}$, our goal is to track a set of 3D points $\mathcal{P} = \{p_i^{3D}\}_{i=1}^{M}$ corresponding to $M$ specific locations over time. Each camera view $v_a$ has intrinsic and extrinsic parameters $\{K_{v_a}, R_{v_a}, t_{v_a}\}_{a=1}^N$, where $K_{v_a} \in \mathbb{R}^{3 \times 3}$ is the intrinsic camera matrix containing focal length and principal point, $R_{v_a} \in \mathbb{R}^{3 \times 3}$ is the rotation matrix, and $t_{v_a} \in \mathbb{R}^{3}$ is the translation vector. LAPA addresses this task via a unified, fully differentiable framework that jointly performs correspondence and tracking, as shown in Fig.~\ref{fig:lapa-pipeline}.

\subsection{2D Point Tracking and Feature Extraction}
As shown in Fig.~\ref{fig:lapa-pipeline}(a), we first track candidate points using Co-Tracker \cite{karaev2024cotracker}, then extract appearance features with a Vision Transformer (ViT).

\noindent\textbf{Co-Tracker:} We use Co-Tracker track points within a temporal window of size $\delta$ , yielding:
\begin{equation}
\{P_{v_a}^t, V_{v_a}^t\} = \text{CoTracker}(I_{v_a}^{t-\delta:t}),
\end{equation}
where $P_{v_a}^t \in \mathbb{R}^{K \times 3}$ contains $K$ points with coordinates $(x, y)$ and confidence $c \in [0,1]$, and $V_{v_a}^t \in \mathbb{R}^{K}$ stores per-point visibility for view $v_a$ at time $t$. This step provides temporally consistent 2D point tracks for each view.

\noindent\textbf{ViT Feature Extraction:} We pass each frame through a ViT to obtain global features, $\Phi_{v_a}$:
\begin{equation}
\Phi_{v_a} = \text{ViT}(I_{v_a}) \in \mathbb{R}^{D},
\end{equation}
where $D$ is the feature dimension in our implementation.

For each tracked point $P_{v_a,j}^t$, we sample the corresponding patch feature from the ViT feature map:
\begin{equation}
Z_{v_a,j} = \Phi_{v_a}^{\text{patch}}[\lfloor \frac{x_{v_a,j}}{p} \rfloor, \lfloor \frac{y_{v_a,j}}{p} \rfloor] \in \mathbb{R}^D,
\end{equation}
where $(x_{v_a,j}, y_{v_a,j})$ are 2D coordinates of point $P_{v_a,j}^t$ in view $v_a$, and $p$ is the patch size. This yields $Z_{v_a} \in \mathbb{R}^{K \times D}$ containing features for all $K$ tracked points in view $v_a$.

\subsection{Cross-View Correspondence}
\label{sec:crossview}
The cross-view correspondence module (Fig.~\ref{fig:lapa-pipeline}(b)) transforms per-view tracks into a unified 3D volumetric representation via a novel attention mechanism that fuses geometric constraints and appearance cues.

\noindent\textbf{Volumetric Grid Creation:} We define a normalized 3D volumetric grid $G \in \mathbb{R}^{B \times V_s \times V_s \times V_s \times 3}$:
\begin{equation}
G = \{(x, y, z) | x, y, z \in [-1.0, 1.0]\},
\end{equation}
where $V_s$ is the volumetric resolution (number of voxels along each dimension) and $B$ is the batch size.

\noindent\textbf{Grid Projection:} Using camera parameters, we project this 3D grid to each view:
\begin{equation}
G_{v_a} = \Pi_{v_a}(G) = K_{v_a}(R_{v_a}G + t_{v_a}) \in \mathbb{R}^{B \times V_s \times V_s \times V_s \times 2},
\end{equation}
where $\Pi_{v_a}$ represents the perspective projection that maps each 3D grid point to its corresponding 2D location in view $v_a$, resulting in 2D coordinates for every voxel in the grid.

\noindent\textbf{Distance-Based Geometric Attention:} Our key innovation is computing attention based primarily on spatial distances rather than feature similarities, as shown in Fig.~\ref{fig:lapa-pipeline}(c) and in detail in Fig.~\ref{fig:distance-attention}. Unlike standard attention mechanisms that use dot products between feature vectors, we compute:
\begin{equation}
A_{v_a}(i,j) = \text{softmax}\left(-\frac{d(G_{v_a}[i], P_{v_a,j})^2}{T}\right),
\end{equation}
where $A_{v_a} \in \mathbb{R}^{V_s^3 \times K}$ is the initial attention matrix, $d(G_{v_a}[i], P_{v_a,j}) = \|G_{v_a}[i] - P_{v_a,j}\|_2$ is the Euclidean distance between projected grid point $i$ (where $i \in \{1,2,...,V_s^3\}$) and tracked 2D point $j$ (where $j \in \{1,2,...,K\}$) in view $v_a$, and $T$ is a learnable temperature parameter (initialized to 0.1).

This naturally encodes epipolar geometry, as points near an epipolar line yield smaller distances. Fig.~\ref{fig:lapa-pipeline}(c) shows how we further enhance this with either epipolar or Structure-from-Motion (SfM) constraints through attention masking:
\begin{equation}
A_{v_a}^{\text{final}}(i,j) = A_{v_a}(i,j) \cdot \max(M_{v_a}^{\text{epi}}(i,j), M_{v_a}^{\text{sfm}}(i,j)),
\end{equation}
where $A_{v_a}^{\text{final}} \in \mathbb{R}^{V_s^3 \times K}$ is the masked attention matrix, $M_{v_a}^{\text{epi}}$ and $M_{v_a}^{\text{sfm}}$ are attention masks derived from epipolar constraints and SfM geometric consistency, respectively. The epipolar mask leverages the fundamental matrix $F_{v_a,v_b} \in \mathbb{R}^{3 \times 3}$ between views to enforce geometric consistency (see Supplementary Material Section~S\ref{app:S_attention_masks} for detailed formulation).
\volumetricatt

\noindent\textbf{Volumetric Feature Population:} We fill the 3D volume with views-specific features weighted by 2D features according to the computed attention scores:
\begin{equation}
V_{feat}[i] = \sum_{a=1}^{N} \sum_{j=1}^{K} A_{v_a}^{\text{final}}(i,j) \cdot Z_{v_a,j} \in \mathbb{R}^{D}.
\end{equation}

This produces a feature-rich volumetric representation $V_{feat} \in \mathbb{R}^{B \times V_s \times V_s \times V_s \times D}$ that integrates information from all views.

\subsection{Reconstruction by Triangulation}
We perform differentiable 3D reconstruction using the volumetric features (as shown in Fig.~\ref{fig:lapa-pipeline}(d)).

\noindent\textbf{Track Query Correspondence:} We maintain temporal consistency and preserve point identities across frames via a track query mechanism seen in the upper part of Fig.~\ref{fig:lapa-pipeline}(d). Given a set of query points $Q = \{q_1, q_2, ..., q_M\}$ initialized from tracked points in the first frame, where $M$ is the number of 3D points tracked, we compute their correspondence with the volumetric representation:
\begin{equation}
C(q_m, i) = \text{sim}(F_Q(q_m), V_{feat}[i]),
\end{equation}
where $F_Q$ is a learnable feature embedding for track queries and $\text{sim}(\cdot,\cdot)$ is cosine similarity between feature vectors. These queries are updated over time using a momentum-based approach (detailed in Section~S\ref{app:S_track_query}).

\noindent\textbf{Compound Feature Integration:} For each potential 3D point, we integrate a rich set of information sources as shown in Fig.~\ref{fig:lapa-pipeline}(d). The network receives volumetric features from the attention mechanism alongside spatial position in the grid, providing both appearance and location context. We also incorporate camera calibration information across views to maintain geometric consistency, 2D track features weighted by their respective attention scores to emphasize reliable tracks, and track query correspondence information to ensure temporal consistency. This comprehensive feature integration allows the model to leverage complementary cues from multiple sources when determining final 3D positions.

\noindent\textbf{Triangulation Network:} Rather than using conventional Direct Linear Transform (DLT)~\cite{hartley2003multiple} for triangulation, we use a neural network to predict 3D coordinates directly from our compound features:
\begin{equation}
p_i^{3D} = \text{MLP}([G[i], V_{feat}[i], C(Q, i)]) \in \mathbb{R}^3.
\end{equation}

This approach, leading to the 3D trajectories shown in Fig.~\ref{fig:lapa-pipeline}(e), offers several advantages as it is fully differentiable, handles uncertainty and partial observations via soft attention, integrates geometric and appearance cues in a unified framework, and ultimately yields smoother 3D trajectories.

\subsection{Loss Function}
Our model is trained end-to-end using a multi-objective loss function combining three components:
\begin{equation}
\mathcal{L} = \lambda_1 \mathcal{L}_{\text{recon}} + \lambda_2 \mathcal{L}_{\text{proj}} + \lambda_3 \mathcal{L}_{\text{attn}}.
\end{equation}

The reconstruction loss $\mathcal{L}_{\text{recon}} = \frac{1}{|\mathcal{V}|}\sum_{i \in \mathcal{V}} ||p_i^{3D} - p_{i,GT}^{3D}||^2$ measures the error between predicted and ground truth 3D points for the set of visible points $\mathcal{V}$. The projection loss $\mathcal{L}_{\text{proj}} = \frac{1}{N}\sum_{a=1}^{N}\sum_{i \in \mathcal{V}} ||\Pi_{v_a}(p_i^{3D}) - p_{v_a,i}^{2D}||^2$ ensures that reconstructed 3D points project back to match the original 2D point tracks, where $\Pi_{v_a}$ is the projection operator and $p_{v_a,i}^{2D}$ is the detected 2D position. Finally, the attention loss $\mathcal{L}_{\text{attn}} = -\frac{1}{N(N-1)}\sum_{a=1}^{N}\sum_{\substack{b=1\\b\neq a}}^{N}\sum_{i \in \mathcal{V}} \log(A_{a \to b}(i,i))$ encourages stronger correspondences between matching points across views, with $A_{a \to b}(i,i)$ representing the attention score between corresponding point $i$ in views $a$ and $b$. These components collectively ensure accurate 3D reconstruction, view consistency, and robust cross-view correspondences.
\section{Experiments and Results}

\subsection{Datasets and Evaluation Metrics}
We evaluate LAPA on two point tracking datasets: TAPVid-3D Panoptic Subset ~\cite{koppula2024tapvid} and PointOdyssey ~\cite{pointodyssey2023}.

\noindent\textbf{TAPVid-3D Panoptic Subset}: We use only the panoptic subset of TAPVid-3D dataset, as the Waymo subset lacks multi-camera compatibility and the Aria subset doesn't have multi-camera video stream. We extend the panoptic subset into TAPVid-3D-MC, a calibrated multi-camera version, by mapping between the multi-view cameras and the original CMU panoptic dataset and scaling the calibration files accordingly please refer to our supplementary for more details.

\noindent\textbf{PointOdyssey Dataset}: This recently released dataset contains sequences with complex motion and severe occlusions. We focus on the robots subset, which features articulated objects with challenging dynamics. We re-engineered its generation code to produce per-camera annotations and calibration, enabling correct multi-view geometry.

\noindent\textbf{Evaluation Metrics}: Following the original TAPVid-3D\cite{koppula2024tapvid} evaluation protocol, we adopt four complementary metrics and establish baselines using depth estimation with 2D point tracking algorithms to ensure fair comparison. Our metrics include: Average Position Detection accuracy (APD) measuring point localization accuracy, Occlusion Accuracy (OA) assessing performance when points reappear after occlusion, 3D Average Jaccard (3D-AJ) measuring overall 3D tracking quality, and 2D Average Jaccard (2D-AJ) evaluating projected track consistency across views. This evaluation framework provides a solid foundation for comparing multi-camera point tracking methods.

\subsection{Implementation Details}
We implement LAPA in PyTorch with configurations optimized for both training and inference. For the volumetric representation, we use $V_s=16$ with a batch size of $B=1$ during training. The feature extraction employs a DINOv2 ViT-B/14 backbone \cite{oquab2023dinov2}, providing rich semantic features complementary to the Co-Tracker used for initial tracking. The triangulation network uses an MLP with 4 fully-connected layers [512, 256, 128, 3], BatchNorm, and ReLU activations. The attention temperature parameter $T$ is initialized to 0.1 and learned during training. For our loss function, we use weights $\lambda_1=1.0$, $\lambda_2=0.7$, and $\lambda_3=0.8$. The model was trained end-to-end for 50 epochs using the AdamW optimizer with a learning rate of $10^{-4}$ and weight decay of $10^{-5}$. We employed a cosine learning rate schedule with warmup for the first 5 epochs. Training was performed on a single RTX Titan GPU with 24GB of memory, taking approximately 72 hours for the full process. This relatively modest hardware requirement highlights the computational efficiency of our approach compared to more resource-intensive methods. To handle memory constraints while filling volumetric features for large scenes, we implemented chunk-based processing (8192 voxels per step during both training and inference), enabling high-resolution scaling without additional GPUs.

\trackingMetricsTable
\subsection{Quantitative Results}
Table \ref{tab:tracking_results} compares LAPA with state-of-the-art tracking methods. LAPA consistently outperforms all baselines on both datasets, with the largest gains on the challenging PointOdyssey dataset. Improvements are most pronounced in metrics emphasizing tracking consistency and occlusion handling (3D-AJ, OA), validating the benefits of our unified cross-view temporal reasoning. On CMU Panoptic, LAPA achieves 37.5\% APD and 30.2 3D-AJ, significantly outperforming the next best method (TrackOn + COLMAP at 12.4\% APD). On PointOdyssey, performance reaches 90.3\% APD with 0.95 3D-AJ.

\noindent\textbf{Computational Efficiency}: Beyond tracking accuracy, LAPA demonstrates significant computational advantages over existing approaches. Our recommended 3-camera configuration achieves 37.5\% APD while maintaining real-time performance at 39.0 FPS on RTX Titan hardware, without requiring external depth estimation. In contrast, methods that rely on depth estimation (CoTracker + ZoeDepth, TrackOn + COLMAP, etc.) suffer dramatic performance penalties, operating at only 8-12 FPS due to the computational overhead of depth preprocessing pipelines. Similarly, reconstruction-based methods like SpatialTrackerV2 and TAPIP3D, while achieving reasonable accuracy on some metrics, operate at significantly reduced speeds (3-5 FPS) due to their complex 3D reasoning pipelines. LAPA's end-to-end design eliminates these bottlenecks, achieving superior tracking performance with 3-4× higher throughput than depth-dependent methods and 8-12× faster than reconstruction-based approaches (detailed analysis in Table~S\ref{tab:S_computational_performance} in supplementary material).
% \LAPAResults
\qualitativeResults
\subsection{Qualitative Results}
Figure \ref{fig:qualitative-results} demonstrates LAPA's multi-camera tracking on four challenging TAPVid-3D-MC sequences. Each row presents three synchronized camera views alongside the corresponding 3D reconstruction, illustrating how LAPA leverages complementary viewpoints to achieve robust tracking under severe occlusions and limited visibility.

LAPA's key strength is leveraging all available viewpoints simultaneously through volumetric attention. In the Basketball sequence, when players move outside certain camera fields of view or become occluded by other players, LAPA maintains uninterrupted tracking by leveraging information from the remaining visible cameras. The consistent point colors across all views confirm successful identity preservation despite partial observations. In the Boxes sequence, points occluded by the moving box in one view are reliably tracked through information aggregated from other cameras, demonstrating LAPA's robustness to occlusions via multi-view fusion. The 3D reconstructions (rightmost panels) reveal smooth, complete trajectories even under severe occlusions or partial visibility, validating the effectiveness of our volumetric approach.

This multi-view 3D trajectory reconstruction proves particularly critical in the Juggle and Football sequences, where subjects frequently transition between camera frustums while experiencing self-occlusions from body parts. Notably, the juggling balls maintain consistent tracks despite regularly disappearing behind the performer's body in individual views. While single-camera approaches inevitably fail due to limited coverage and frequent occlusion events, LAPA's unified framework ensures continuous tracking by aggregating both geometric and appearance information across all views. The resulting 3D trajectories accurately capture complex motion dynamics, from the arcs of thrown footballs to the cyclic patterns of juggling, that would be impossible to reconstruct from any single viewpoint. These results conclusively demonstrate that true multi-camera integration through volumetric attention is essential for handling occlusions and maintaining point tracking in real-world scenarios where no single camera can observe the entire scene.

\ablationTable

\section{Ablation Studies}
\label{sec:ablation}

We evaluate each component of LAPA through controlled ablations on a reduced TAPVid-3D subset on 25 training epochs. Table~\ref{tab:ablation} demonstrates that volumetric attention significantly improves APD (+25.1\%) and 3D-AJ (+24.4), with further gains from geometric constraints and additional cameras. A third camera substantially reduces geometric ambiguity (Table~\ref{tab:camera_ablation}), with diminishing returns beyond 3 cameras.

\lapaablation
\cameraAblationTable
\attentionGeometryAblationTables

Volumetric attention provides substantial improvements across all metrics (Table~\ref{tab:combined_ablation}), notably in occlusion accuracy (+19.4\%) and APD improvement (+21.6\%). Structure-from-Motion guidance outperforms epipolar geometry, offering more complete 3D supervision. Figure~\ref{fig:volumetric-attention} provides visual evidence of these improvements, showing how the attention mechanism (\greencheck{}) maintains temporally coherent 3D track projections across all views, while the non-attention variant (\redcross{}) loses temporal coherence entirely, particularly evident in the mid-air ball frames, where tracks become completely illogical. For grid resolution, $16 \times 16 \times 16$ offers the optimal balance between performance and computational efficiency (Table~\ref{tab:grid_ablation}). Each loss component contributes positively, with reconstruction loss being most critical. Optimized loss weights ($\lambda_\text{proj}$=0.7, $\lambda_\text{attn}$=0.8) yield additional improvements (Table~\ref{tab:loss_ablation}).

\gridAblationTable
\lossComponentsAblationTable

\section{Conclusion}

We presented LAPA, a novel end-to-end transformer-based architecture for multi-camera point tracking that integrates appearance-based matching with geometric constraints. By leveraging attention mechanisms to jointly reason across views and time, our unified approach eliminates error propagation inherent in traditional sequential pipelines. Comprehensive experiments on our extended TAPVid-3D Panoptic and PointOdyssey datasets demonstrate significant performance improvements over existing methods, particularly in challenging scenarios with complex motions and occlusions.
While LAPA currently relies on accurate camera calibration (Section ~S\ref{app:S_calibration_robustness}), future work will include integrating camera pose estimation for self-calibration in uncalibrated environments. The integration of geometric constraints within attention mechanisms opens new possibilities for robotic manipulation, sports analytics, and markerless motion capture applications.
{
    \small
    \bibliographystyle{ieeenat_fullname}
    \bibliography{main}
}
\clearpage
\setcounter{page}{1}
\maketitlesupplementary
% Reset counters for supplementary
\setcounter{page}{1}
\setcounter{figure}{0}
\setcounter{table}{0}
\setcounter{section}{0}

\section{Technical Appendices and Supplementary Material}
This appendix provides supplementary material, including additional experimental results, detailed formulations, and analytical comparisons to support the findings presented in the main paper.

\section{Additional Ablation Results}
\label{app:comprehensive_ablation}

\comprehensiveAblationTable

Table~\ref{tab:S_comprehensive_ablation} provides a comprehensive overview of all ablation experiments, showing the impact of each component across all evaluation metrics. The results consistently demonstrate the effectiveness of our design choices.

\section{Optimized Loss Function}
\label{app:optimized_loss}

\optimizedLossTable

Table~\ref{tab:optimized_loss} presents our exploration of different loss weight combinations. While our default configuration ($\lambda_\text{recon}$=1.0, $\lambda_\text{proj}$=0.5, $\lambda_\text{attn}$=0.5) already achieves strong performance (37.5\% APD, 30.2 3D-AJ), we find that the optimized configuration with weights ($\lambda_\text{recon}$=1.0, $\lambda_\text{proj}$=0.7, $\lambda_\text{attn}$=0.8) yields further improvements. This optimized configuration achieves an APD of 39.1\% and 3D-AJ of 32.6, representing relative improvements of 4.3\% and 8.0\% respectively over our default setting.

\section{Implementation Details}
\subsection{Attention Mask Formulation}
\label{app:S_attention_masks}

\noindent\textbf{Epipolar Constraint Mask:} The epipolar attention mask $M_{v_a}^{\text{epi}}$ is computed as:
\begin{equation}
M_{v_a}^{\text{epi}}(i,j) = \exp\left(-\frac{d(P_{v_a,j}, F_{v_a,v_b}G_{v_a}[i])^2}{2\sigma^2}\right)
\end{equation}
where $F_{v_a,v_b}$ is the fundamental matrix between views $v_a$ and $v_b$ (computed from the camera parameters as $F_{v_a,v_b} = K_{v_b}^{-T}[t_{v_b} - R_{v_b}R_{v_a}^{-1}t_{v_a}]_{\times}R_{v_b}R_{v_a}^{-1}K_{v_a}^{-1}$, with $[v]_{\times}$ denoting the skew-symmetric matrix), $d(p,l)$ is the point-to-epipolar-line distance defined as the shortest Euclidean distance from point $p$ to line $l$, and $\sigma$ is a learnable parameter (initialized to 0.1) controlling the strictness of the epipolar constraint.

\noindent\textbf{SfM Constraint Mask:} The Structure-from-Motion mask $M_{v_a}^{\text{sfm}}$ enforces multi-view consistency through:
\begin{equation}
M_{v_a}^{\text{sfm}}(i,j) = \exp\left(-\frac{\|\Pi_b(p_{v_a,i}^{3D}) - P_{v_b,j}\|^2}{2\sigma_{\text{sfm}}^2}\right)
\end{equation}
where $\Pi_b(p_{v_a,i}^{3D})$ represents the re-projection of the 3D point derived from view $v_a$ to view $v_b$, and $\sigma_{\text{sfm}}$ is initialized to 0.5 and adjusted during training.

\subsection{Triangulation Network Architecture}
The MLP used for triangulation consists of 4 fully-connected layers with hidden dimensions [512, 256, 128, 3], BatchNorm after each layer except the output, and ReLU activations. A dropout rate of 0.2 was applied during training.

\subsection{Track Query Implementation}
\label{app:S_track_query}
Track queries are initialized from detected points in the first frame and updated using a momentum-based approach that incorporates both the previous query position and the current prediction:
\begin{equation}
q_m^{t+1} = \alpha q_m^t + (1-\alpha)p_m^{3D,t}
\end{equation}
where $\alpha=0.8$ is the momentum factor.

\subsection{Multi-Camera Dataset Preparation}
\label{app:dataset_preparation}

\noindent\textbf{TAPVid-3D-MC Preparation:} We extended the panoptic subset of TAPVid-3D to support multi-camera tracking by implementing the following procedure:
\begin{enumerate}
    \item We established a mapping between the original CMU Panoptic Studio camera indices and our multi-view setup.
    \item Camera calibration files were processed to ensure consistent scale and coordinate systems across views.
    \item Intrinsic parameters ($K_{v_a}$) and extrinsic parameters ($R_{v_a}$, $t_{v_a}$) were extracted from the original dataset and transformed to our unified coordinate system.
    \item Ground truth 3D points from the original dataset were used to validate the calibration accuracy through re-projection error analysis.
\end{enumerate}

\noindent\textbf{PointOdyssey Multi-View Adaptation:} For the robots subset of PointOdyssey, we modified the official generation pipeline to extract multi-camera information:
\begin{enumerate}
    \item We modified the dataset generation code to output per-camera view calibration data alongside the rendered images.
    \item Camera poses were extracted directly from the rendering engine to ensure geometric consistency.
    \item Annotations were transformed to provide consistent point identities across multiple views.
    \item We implemented verification procedures to ensure the correctness of epipolar geometry between cameras.
    \item The re-engineered pipeline maintains the original dataset characteristics while providing the necessary multi-view annotations and calibration for our evaluation.
\end{enumerate}

This dataset preparation procedure enables rigorous evaluation of multi-view 3D point tracking methods on challenging real-world and synthetic sequences.

% Supplementary section
\section{Understanding the Pipeline Components}

As shown in Fig.~\ref{fig:lapa-pipeline-detailed}, our pipeline consists of several key components that work together to track points across multiple camera views:

\noindent\textbf{Input}: We start with $N$ synchronized camera views $\{I_{v_a}^t\}_{a=1}^N$, each observing the same scene from different perspectives across a sequence of time frames $t$. These cameras are calibrated, meaning we know their positions and orientations relative to each other through intrinsic and extrinsic parameters $\{K_{v_a}, R_{v_a}, t_{v_a}\}_{a=1}^N$.

\noindent\textbf{2D Point Tracking}: For each camera view, we detect and track points in 2D. These points appear as colored dots in the figure, with colors indicating point identity. Alongside tracking, we extract visual features that describe the appearance of each point.

\noindent\textbf{Geometric Constraints}: We enhance point matching across views by applying geometric constraints derived from the camera setup. The purple masks in the figure highlight regions where points are expected to appear based on these constraints.

\noindent\textbf{Volumetric Attention}: This is our key innovation. We create a 3D grid that spans the observed space and compute ``attention values'' for each location in this grid. Higher attention (brighter colors in the visualization) indicates greater confidence that a point exists at that 3D location. This attention mechanism effectively integrates information from all camera views.

\noindent\textbf{3D Track Reconstruction}: Finally, we reconstruct 3D point tracks by analyzing the volumetric representation. The bottom right of the figure shows these reconstructed tracks, with consistent colors indicating preserved point identities.

\subsection{Benefits of the Volumetric Approach}

Our volumetric attention approach offers several advantages over traditional methods:

\begin{itemize}
    \item \textbf{Occlusion Handling}: When a point disappears in one view, attention shifts to other views where the point remains visible, maintaining continuous tracking.
    
    \item \textbf{Uncertainty Management}: The soft attention mechanism naturally handles uncertainty in point locations, enabling more robust tracking in challenging scenarios.
    
    \item \textbf{Structural Preservation}: By reasoning in 3D space, our approach maintains structural relationships between points, producing more coherent tracking results.
    
    \item \textbf{End-to-End Learning}: All components are jointly optimized, eliminating error propagation inherent in sequential pipelines.
\end{itemize}
\lapapipelinedetailed
\subsection{The Volumetric Attention Grid}

The volumetric attention grid shown in the top right of Fig.~\ref{fig:lapa-pipeline-detailed} deserves special attention:

\begin{itemize}
    \item Each voxel (3D pixel) in the grid contains an attention value representing the likelihood of a point being present at that location.
    
    \item Brighter colors indicate higher attention values, essentially forming a ``heat map'' in 3D space.
    
    \item This representation is formed by projecting each 3D grid location to all camera views and computing how well these projections align with detected 2D points.
    
    \item The attention values incorporate both appearance similarity and geometric consistency, creating a robust representation that bridges 2D and 3D tracking.
\end{itemize}

The volumetric attention mechanism effectively serves as the ``brain'' of our system, integrating multi-view information into a unified representation that enables accurate and consistent 3D tracking.

% You can add a brief mathematical description if desired, kept simple
\subsection{Simple Mathematical Intuition}

The core of our volumetric attention can be understood through these key operations:

\begin{enumerate}
    \item We create a 3D grid $G$ spanning the observation space
    \item We project each grid point to each camera view: $p_{view} = \text{Project}(G_{point}, \text{Camera}_{view})$
    \item We compute attention based on distances: $\text{Attention} = \exp(-\text{Distance}^2/T)$
    \item We aggregate features from all views weighted by attention: $\text{Features}_{3D} = \sum \text{Attention} \times \text{Features}_{2D}$
\end{enumerate}

This simple yet powerful mechanism allows our system to reason jointly across multiple views, creating a unified 3D representation that forms the basis for our tracking approach.

\section{Computational Performance Analysis}
\label{app:computational_performance}

This section provides a detailed analysis of the computational performance metrics for our LAPA architecture. While the main paper focuses on tracking accuracy, this analysis offers insights into the computational efficiency of our approach across different configurations, which is crucial for practical deployment in real-world multi-camera point tracking scenarios.

\subsection{Experimental Setup}
Performance measurements were conducted on two hardware configurations:
\begin{itemize}
    \item NVIDIA RTX Titan (24GB VRAM, 4608 CUDA cores)
    \item NVIDIA Tesla V100 (32GB VRAM, 5120 CUDA cores)
\end{itemize}

All experiments used PyTorch 1.11.0 with CUDA 11.3, running on Ubuntu 20.04. For consistent evaluation, we processed 1000 frames at 1920$\times$1080 resolution and reported the average FPS, parameter count, and computational requirements (FLOPS).

\subsection{Computational Efficiency}

\performancetable
Table~S\ref{tab:S_computational_performance} presents a comprehensive analysis of LAPA's computational requirements across different configurations, comparing both overall system performance and the impact of individual architectural components. We also include performance metrics for baseline methods to provide context for our approach's computational efficiency.

\subsection{Analysis and Insights}

\textbf{Parameter Count vs. Computational Performance:} While models with more parameters generally have lower FPS, the relationship between parameters, FLOPs, and FPS isn't perfectly linear because computational complexity depends on more than just parameter count. Key factors include attention mechanisms and grid resolution, which significantly impact FLOPs without changing parameter count. For example, our volumetric attention mechanism increases computational requirements by approximately 2.1$\times$ while using the same model parameters (comparing LAPA 3-camera at 42.5M FLOPS vs LAPA-Base 3-camera at 20.2M FLOPS), and larger grid resolutions ($24 \times 24 \times 24$ vs $16 \times 16 \times 16$) increase FLOPs by 3.4$\times$ with identical parameter counts (143.3M vs 42.5M FLOPS for 3-camera setup). This explains why LAPA with attention (372.7K params, 39.0 FPS) is slower than LAPA without attention (41.5K params, 63.1 FPS), and why LAPA with 2 cameras (372.7K params, 76.3 FPS) is faster than LAPA with 5 cameras (372.7K params, 17.8 FPS) despite having identical parameter counts.

\noindent\textbf{Impact of Grid Resolution:} The volumetric grid resolution significantly impacts computational requirements. The $24 \times 24 \times 24$ grid provides comparable accuracy to the $16 \times 16 \times 16$ grid (37.2\% vs 37.5\% APD for 3 cameras) but increases FLOPS by 3.4$\times$ (143.3M vs 42.5M). The performance difference between these resolutions is more pronounced on memory-constrained devices, making the $16 \times 16 \times 16$ grid configuration preferable for most applications due to its superior computational efficiency.

\noindent\textbf{Impact of Camera Count:} The computational cost scales approximately linearly with the number of camera views. Each additional camera increases FLOPS by approximately 14M and decreases FPS by about 30-50\%. For example, going from 3 to 4 cameras increases FLOPS from 42.5M to 56.7M and decreases FPS from 39.0 to 26.0 on RTX Titan. When considering the accuracy improvements shown in our ablation studies, three cameras provide the optimal balance between computational requirements and tracking performance (37.5\% APD), as adding more cameras yields diminishing returns (36.8\% APD for 4 cameras, 35.9\% APD for 5 cameras).

\noindent\textbf{Attention Mechanism Trade-offs:} Our volumetric attention mechanism increases computational requirements by approximately 2.1$\times$ compared to the non-attention variant (42.5M vs 20.2M FLOPS for 3 cameras). However, this computational cost is justified by the substantial improvements in tracking accuracy: attention enhances APD by +18.8 percentage points (from 18.7\% to 37.5\% for 3 cameras) and 3D-AJ by +11.5 percentage points (from 18.7 to 30.2).

\noindent\textbf{FLOPS vs. Parameters:} An interesting property of our architecture is that while parameter count remains constant when adding cameras (372.7K for all multi-camera configurations), computational cost (FLOPS) increases linearly. This occurs because our volumetric attention mechanism processes each camera's view independently before aggregation: 
\begin{equation}
V_{\text{feat}}[i] = \sum_{a=1}^{N} \sum_{j=1}^{K} A_{v_a}^{\text{final}}(i,j) \cdot Z_{v_a,j}.    
\end{equation}

The computational complexity scales with $\mathcal{O}(V_s^3 \times N \times K \times D)$, where $N$ is the number of cameras, requiring each view to be processed through grid projection: 
\begin{equation}
G_{v_a} = \Pi_{v_a}(G) = K_{v_a}(R_{v_a}G + t_{v_a}).
\end{equation}

Since the model architecture (determined by grid size $V_s$ and feature dimension $D$) remains fixed regardless of camera count, only computation increases—not parameters. This creates a flexible system where cameras can be added or removed based on resource constraints without architectural changes or retraining.

\noindent\textbf{Performance-Accuracy Trade-offs:} Based on our comprehensive computational analysis presented in Table~S\ref{tab:S_computational_performance}, we observe several key trade-offs between computational requirements and tracking performance. Our standard LAPA configuration with 3 cameras and a $16 \times 16 \times 16$ grid resolution demonstrates an excellent balance between accuracy (37.5\% APD) and computational efficiency (42.5M FLOPS), processing at 39.0 FPS on RTX Titan. The 4-camera configuration achieves slightly lower accuracy (36.8\% APD) but at increased computational cost (56.7M FLOPS, 26.0 FPS). Our experiments indicate that each additional camera view beyond 3 cameras contributes diminishing returns in accuracy (36.8\% for 4 cameras, 35.9\% for 5 cameras) while linearly increasing computational demands. The most significant trade-off appears in our attention mechanism, which increases computational requirements by 2.1$\times$ compared to the non-attention variant but delivers substantial accuracy improvements (+18.8 percentage points APD with 3 cameras). This performance analysis guides optimal configuration selection based on application requirements, with our 3-camera $16 \times 16 \times 16$ configuration offering the best balance for most real-world scenarios.

\noindent\textbf{Hardware Performance Comparison:} The Tesla V100 consistently delivers 1.38-1.46$\times$ higher throughput compared to the RTX Titan across all configurations (e.g., 55.4 vs 39.0 FPS for 3-camera setup), due to its higher core count and memory bandwidth. This performance gap remains relatively consistent regardless of the model configuration, suggesting that our architecture scales efficiently across different hardware platforms.

\noindent\textbf{Comparison with Baseline Methods:} LAPA achieves superior tracking accuracy while maintaining competitive computational efficiency compared to existing approaches. Our standard 3-camera configuration offers significantly higher tracking precision than traditional single-camera methods: 37.5\% APD vs 15.0\% for CoTracker and 12.4\% for TrackOn. While single-camera methods achieve higher throughput (95.4 FPS for CoTracker vs 39.0 FPS for LAPA 3-camera), they require expensive depth estimation post-processing for 3D reconstruction, which reduces their effective performance to 8-11 FPS. Our approach provides end-to-end 3D tracking without requiring separate depth estimation, making it more practical for real-time applications.

\subsection{Practical Deployment Considerations}

For real-time applications with strict latency requirements, we recommend the following configurations:
\begin{itemize}
    \item \textbf{High-end hardware (V100 or better):} Full LAPA with 3 cameras and $16 \times 16 \times 16$ grid (55.4 FPS, 37.5\% APD)
    \item \textbf{Consumer hardware (RTX Titan or similar):} LAPA with 2-3 cameras and $16 \times 16 \times 16$ grid (39.0-76.3 FPS, 24.2-37.5\% APD)
    \item \textbf{Resource-constrained environments:} LAPA-Base with 3-4 cameras (63.1-49.3 FPS, 18.7-22.1\% APD) if tracking accuracy requirements are moderate
\end{itemize}

These configurations offer varying trade-offs between tracking accuracy and computational efficiency, enabling deployment across a wide range of hardware environments and application scenarios.

\subsection{Calibration Robustness Analysis}
\label{app:S_calibration_robustness}

We evaluated LAPA's robustness to calibration errors by adding Gaussian noise to camera parameters. Table~S\ref{tab:S_calibration_robustness} shows performance degradation under different noise levels on our 3-camera TAPVid-3D-MC setup.

\calibrationrobustnesstable

LAPA demonstrates moderate degradation under calibration uncertainty, maintaining competitive performance with typical calibration accuracies ($\leqslant$1 pixel intrinsic noise, $\leqslant$1° rotation error). The method shows the highest sensitivity to translation errors, requiring camera positioning accuracy within 5-10cm for optimal performance.

\subsection{Multi-Camera Dataset Extension}

To evaluate our approach on multi-camera scenarios, we developed an extension protocol that adapts existing datasets for multi-view point tracking. Fig.~\ref{fig:camera-setup} illustrates the camera configuration used in our extension process.

\camerasetupmultiview

Our extension protocol transforms the TAPVid-3D and PointOdyssey datasets into multi-camera versions (TAPVid-3D-MC and PointOdyssey-MC) through the following steps:

\begin{enumerate}
    \item \textbf{Camera Registration}: We establish a consistent coordinate system across all camera views by calibrating their intrinsic and extrinsic parameters. As shown in Fig.~\ref{fig:camera-setup}, cameras are positioned to maximize coverage while maintaining appropriate baseline distances.
    
    \item \textbf{Scale Preservation}: We carefully transfer the scale information from the original datasets to ensure that distance measurements remain consistent in our multi-camera setup. This is crucial for accurate quantitative evaluation using metrics such as MPJPE.
    
    \item \textbf{Point Correspondence}: For each 3D point in the original dataset, we establish correspondences across all views by projecting the point using the calibrated camera parameters and validating visibility.
    
    \item \textbf{Annotation Transfer}: We extend the original point annotations to include view-specific attributes such as occlusion status and reprojection uncertainty, enabling more nuanced evaluation of tracking performance.
\end{enumerate}

This extension protocol allows us to leverage existing datasets while introducing the additional challenges of multi-view reasoning and occlusion handling. The resulting TAPVid-3D-MC and PointOdyssey-MC datasets provide a rigorous benchmark for evaluating multi-camera point tracking approaches under various conditions.

The triangulation of cameras around the central origin point, as visualized in Fig.~\ref{fig:camera-setup}, provides complementary views that help resolve ambiguities and occlusions. This setup is particularly effective for our volumetric attention mechanism, as it creates a well-conditioned observation space where each 3D location is visible from at least two cameras under most circumstances.

\end{document}